\documentclass[]{easychair}     

\usepackage{stmaryrd}
\usepackage{verbatim}

\newdimen\proofrulebreadth \proofrulebreadth=.05em
\newdimen\proofdotseparation \proofdotseparation=1.25ex
\newdimen\proofrulebaseline \proofrulebaseline=2ex
\newcount\proofdotnumber \proofdotnumber=3
\let\then\relax
\def\hfi{\hskip0pt plus.0001fil}
\mathchardef\squigto="3A3B
\newif\ifinsideprooftree\insideprooftreefalse
\newif\ifonleftofproofrule\onleftofproofrulefalse
\newif\ifproofdots\proofdotsfalse
\newif\ifdoubleproof\doubleprooffalse
\let\wereinproofbit\relax
\newdimen\shortenproofleft
\newdimen\shortenproofright
\newdimen\proofbelowshift
\newbox\proofabove
\newbox\proofbelow
\newbox\proofrulename
\def\shiftproofbelow{\let\next\relax\afterassignment\setshiftproofbelow\dimen0 }
\def\shiftproofbelowneg{\def\next{\multiply\dimen0 by-1 }%
\afterassignment\setshiftproofbelow\dimen0 }
\def\setshiftproofbelow{\next\proofbelowshift=\dimen0 }
\def\setproofrulebreadth{\proofrulebreadth}

\def\prooftree{
\ifnum  \lastpenalty=1
\then   \unpenalty
\else   \onleftofproofrulefalse
\fi
\ifonleftofproofrule
\else   \ifinsideprooftree
        \then   \hskip.5em plus1fil
        \fi
\fi
\bgroup
\setbox\proofbelow=\hbox{}\setbox\proofrulename=\hbox{}%
\let\justifies\proofover\let\leadsto\proofoverdots\let\Justifies\proofoverdbl
\let\using\proofusing\let\[\prooftree
\ifinsideprooftree\let\]\endprooftree\fi
\proofdotsfalse\doubleprooffalse
\let\thickness\setproofrulebreadth
\let\shiftright\shiftproofbelow \let\shift\shiftproofbelow
\let\shiftleft\shiftproofbelowneg
\let\ifwasinsideprooftree\ifinsideprooftree
\insideprooftreetrue
\setbox\proofabove=\hbox\bgroup$\displaystyle 
\let\wereinproofbit\prooftree
\shortenproofleft=0pt \shortenproofright=0pt \proofbelowshift=0pt
\onleftofproofruletrue\penalty1
}

\def\eproofbit{
\ifx    \wereinproofbit\prooftree
\then   \ifcase \lastpenalty
        \then   \shortenproofright=0pt  
        \or     \unpenalty\hfil         
        \or     \unpenalty\unskip       
        \else   \shortenproofright=0pt  
        \fi
\fi
\global\dimen0=\shortenproofleft
\global\dimen1=\shortenproofright
\global\dimen2=\proofrulebreadth
\global\dimen3=\proofbelowshift
\global\dimen4=\proofdotseparation
\global\count255=\proofdotnumber
$\egroup  
\shortenproofleft=\dimen0
\shortenproofright=\dimen1
\proofrulebreadth=\dimen2
\proofbelowshift=\dimen3
\proofdotseparation=\dimen4
\proofdotnumber=\count255
}

\def\proofover{
\eproofbit 
\setbox\proofbelow=\hbox\bgroup 
\let\wereinproofbit\proofover
$\displaystyle
}%
\def\proofoverdbl{
\eproofbit 
\doubleprooftrue
\setbox\proofbelow=\hbox\bgroup 
\let\wereinproofbit\proofoverdbl
$\displaystyle
}%
\def\proofoverdots{
\eproofbit 
\proofdotstrue
\setbox\proofbelow=\hbox\bgroup 
\let\wereinproofbit\proofoverdots
$\displaystyle
}%
\def\proofusing{
\eproofbit 
\setbox\proofrulename=\hbox\bgroup 
\let\wereinproofbit\proofusing
\kern0.3em$
}

\def\endprooftree{
\eproofbit 
  \dimen5 =0pt
\dimen0=\wd\proofabove \advance\dimen0-\shortenproofleft
\advance\dimen0-\shortenproofright
\dimen1=.5\dimen0 \advance\dimen1-.5\wd\proofbelow
\dimen4=\dimen1
\advance\dimen1\proofbelowshift \advance\dimen4-\proofbelowshift
\ifdim  \dimen1<0pt
\then   \advance\shortenproofleft\dimen1
        \advance\dimen0-\dimen1
        \dimen1=0pt
        \ifdim  \shortenproofleft<0pt
        \then   \setbox\proofabove=\hbox{%
                        \kern-\shortenproofleft\unhbox\proofabove}%
                \shortenproofleft=0pt
        \fi
\fi
\ifdim  \dimen4<0pt
\then   \advance\shortenproofright\dimen4
        \advance\dimen0-\dimen4
        \dimen4=0pt
\fi
\ifdim  \shortenproofright<\wd\proofrulename
\then   \shortenproofright=\wd\proofrulename
\fi
\dimen2=\shortenproofleft \advance\dimen2 by\dimen1
\dimen3=\shortenproofright\advance\dimen3 by\dimen4
\ifproofdots
\then
        \dimen6=\shortenproofleft \advance\dimen6 .5\dimen0
        \setbox1=\vbox to\proofdotseparation{\vss\hbox{$\cdot$}\vss}%
        \setbox0=\hbox{%
                \advance\dimen6-.5\wd1
                \kern\dimen6
                $\vcenter to\proofdotnumber\proofdotseparation
                        {\leaders\box1\vfill}$%
                \unhbox\proofrulename}%
\else   \dimen6=\fontdimen22\the\textfont2 
        \dimen7=\dimen6
        \advance\dimen6by.5\proofrulebreadth
        \advance\dimen7by-.5\proofrulebreadth
        \setbox0=\hbox{%
                \kern\shortenproofleft
                \ifdoubleproof
                \then   \hbox to\dimen0{%
                        $\mathsurround0pt\mathord=\mkern-6mu%
                        \cleaders\hbox{$\mkern-2mu=\mkern-2mu$}\hfill
                        \mkern-6mu\mathord=$}%
                \else   \vrule height\dimen6 depth-\dimen7 width\dimen0
                \fi
                \unhbox\proofrulename}%
        \ht0=\dimen6 \dp0=-\dimen7
\fi
\let\doll\relax
\ifwasinsideprooftree
\then   \let\VBOX\vbox
\else   \ifmmode\else$\let\doll=$\fi
        \let\VBOX\vcenter
\fi
\VBOX   {\baselineskip\proofrulebaseline \lineskip.2ex
        \expandafter\lineskiplimit\ifproofdots0ex\else-0.6ex\fi
        \hbox   spread\dimen5   {\hfi\unhbox\proofabove\hfi}%
        \hbox{\box0}%
        \hbox   {\kern\dimen2 \box\proofbelow}}\doll%
\global\dimen2=\dimen2
\global\dimen3=\dimen3
\egroup 
\ifonleftofproofrule
\then   \shortenproofleft=\dimen2
\fi
\shortenproofright=\dimen3
\onleftofproofrulefalse
\ifinsideprooftree
\then   \hskip.5em plus 1fil \penalty2
\fi
}

    {\end{array}$$}
\newcommand{\ignore}[1]{}

\newcommand{\typearrow}{\shortrightarrow}

\newtheorem{theorem}{\rm THEOREM}[section]

\newtheorem{lemma}[theorem]{\rm LEMMA}
\newtheorem{proposition}[theorem]{\rm PROP.}
\newtheorem{definition}[theorem]{\rm DEF.}

\begin{document}

\title{Quantified Conditional Logics are Fragments of HOL\thanks{This work has been presented at the conference on Non-classical Modal and Predicate Logics 2011, Guangzhou (Canton), China, 5-9 December 2011.}}

\titlerunning{Quantified Conditional Logics are Fragments of HOL}

\author{
	     Christoph Benzm\"uller \\ Free University Berlin \\ \url{c.benzmueller@googlemail.com}\and	
             Valerio Genovese \\ University of Luxembourg \\ \url{genovese@di.unito.it}
        }

\authorrunning{C. Benzm\"uller and V. Genovese} 




\maketitle

\begin{abstract}
A semantic embedding of quantified conditional logic in classical higher-order logic is presented.
\end{abstract}

\section{Introduction}




A semantic embedding of propositional conditional logic in classical
higher-order logic HOL (Church's type theory) has been presented in
\cite{R52}. This embedding exploits the natural correspondence between
selection function semantics for conditional logics \cite{Sta68} and
HOL.
In fact, selection function semantics can be seen as an higher-order extension of well-known
Kripke semantics for modal logic and cannot be naturally embedded into first-order logic. 

In this paper we extend the embedding in \cite{R52} to also include
quantification over propositions and individuals. This
embedding of quantified conditional logic in HOL is sound and complete. 

\section{Quantified Conditional Logics}
We extend propositional conditional logics with quantification over propositional variables and over individuals of a first-order domain.
Below, we only consider \emph{constant domains}, i.e., every possible world has the same domain.

Let $\mathcal{IV}$ be a set of first-order (individual) variables,
   $\mathcal{PV}$ a set of propositional variables, and $\mathcal{SYM}$
   a set of predicate symbols of any arity.   Formulas of quantified
  conditional logic are given by the following grammar (where $X^i\in\mathcal{IV}, P\in\mathcal{PV}, k\in\mathcal{SYM}$):
$$ \varphi,\psi ::= P \mid k(X^1,\ldots,X^n) \mid \neg \varphi \mid \varphi \vee \psi \mid \forall X. \varphi \mid \forall P. \varphi \mid \varphi \Rightarrow \psi $$

From the selected set of primitive connectives, other logical
connectives can be introduced as abbreviations: for example,
$\varphi\wedge\psi$, $\varphi \rightarrow \psi$ (material
implication), and $\exists X. \varphi$ abbreviate $\neg (\neg \varphi \vee
\neg \psi)$, $\neg \varphi \vee \psi$ and $\neg\forall X. \neg \varphi$ etc.
Syntactically, quantified conditional logics can be seen as a generalization of
quantified multimodal logic where the index of modality $\Rightarrow$ is a
formula of the same language. For instance, in $(\varphi \Rightarrow \psi)
\Rightarrow \delta$ the subformula $\varphi \Rightarrow \psi$ is the index of the
second occurrence of $\Rightarrow$.

Regarding semantics, many different formalizations have been proposed
(see \cite{Nut80}), here we focus on the \emph{selection function
  semantics} \cite{Chellas80}, which is based on possible world
structures and has been successfully used in \cite{Olivetti07} to
develop proof methods for some conditional logics.  We adapt selection
function semantics for quantified conditional logics.

%
An \emph{interpretation} is a structure $\mathcal{M} = \langle S,f,D,Q,I \rangle$ where, $S$ is a set of possible items called worlds, $f: S \times 2^{S} \mapsto 2^{S}$ is the selection function, $D$ is a non-empty set of \emph{individuals} (the first-order domain), $Q$ is a non-empty collection of subsets of $W$ (the propositional domain), and  $I$ is a classical interpretation function where for each n-ary predicate
symbol $k$, $I(k,w) \subseteq D^n$.
%

A \emph{variable assignment}  $g = (g^{iv},g^{pv})$ is a pair of maps where,
$g^{iv}: \mathcal{IV} \mapsto D$ maps each individual variable in $\mathcal{IV}$ to an object in $D$, and $g^{pv}: $ maps each propositional variable in $\mathcal{PV}$ to a set of worlds in $Q$.
%

\emph{Satisfiability} of a formula $\varphi$ for an interpretation $\mathcal{M} = \langle S,f,D,Q,I \rangle$, a world $s \in S$, and a variable assignment
$g = (g^{iv},g^{pv})$ is denoted as $M,g,s \models \varphi$ and defined as follows, where $[a/Z]g$ denote the assignment
identical to $g$ except that $([a/Z]g)(Z) = a$:

$M,g,s \models k(X^1,\ldots,X^n)$ if and only if $\langle g^{iv}(X^1), \ldots, g^{iv}(X^n) \rangle \in I(k,w)$

$M,g,s \models P$ if and only if $s \in g^{pv}(P)$

$M,g,s \models \neg \varphi$ if and only if $M,g,s \not\models \varphi$ (that is, not $M,g,s \models \varphi$ )

$M,g,s \models \varphi \vee \psi$ if and only if $M,g,s \models \varphi$ or $M,g,s \models \psi$

$M,g,s \models \forall X. \varphi$ if and only if $M,([d/X]g^{iv},g^{pv}), s \models \varphi$ for all $d \in D$

$M,g,s \models \forall P. \varphi$ if and only if $M,(g^{iv},[p/P]g^{pv}),s \models \varphi$ for all $p \in Q$

$M,g,s \models \varphi \Rightarrow \psi$ if and only if $M,g,t \models \psi$ for all $t \in S$ such that $t \in f(s,[\varphi])$
where $[\varphi] = \{u \mid M,g,u \models \varphi\}$

  An interpretation $\mathcal{M} = \langle S,f,D,Q, I \rangle$ is a \emph{model}
  if for every variable assignment $g$ and every formula $\varphi$,
  the set of worlds $\{s\in S \mid M,g,s \models \varphi\}$ is a
  member of $Q$.
  As usual, a conditional formula $\varphi$ is \emph{valid in a model}
  $\mathcal{M}=\langle S,f,D,Q, I \rangle$, denoted with $\mathcal{M}
  \models \varphi$, if and only if for all worlds $s \in S$ and variable assignments $g$ holds $\mathcal{M},g,s \models
  \varphi$. A formula $\varphi$ is a \emph{valid}, denoted $\models
  \varphi$, if and only if it is valid in every model.

$f$ is defined to take $[\varphi]$ (called the \emph{proof set} of $\varphi$ w.r.t. a given model $\mathcal{M}$) instead
of $\varphi$. This approach has the consequence of forcing the so-called \emph{normality} property: given a model $\mathcal{M}$, 
if $\varphi$ and $\varphi'$ are equivalent (i.e., they are satisfied in the same set of worlds), then they index the same formulas w.r.t. to the $\Rightarrow$ modality.
The axiomatic counterpart of the normality condition is given by the
rule (RCEA)
\begin{center}
\begin{prooftree}
  \varphi \leftrightarrow \varphi' \justifies (\varphi \Rightarrow
  \psi) \leftrightarrow (\varphi' \Rightarrow \psi) \using (RCEA)
\end{prooftree}
\end{center}
Moreover, it can be easily shown that the above semantics forces also
the following rules to hold:
\begin{center}
\begin{prooftree}
  (\varphi_1 \wedge \ldots \wedge \varphi_n) \leftrightarrow \psi
  \justifies (\varphi_0 \Rightarrow \varphi_1 \wedge \ldots \wedge
  \varphi_0 \Rightarrow \varphi_n) \rightarrow (\varphi_0 \Rightarrow
  \psi) \using (RCK)
\end{prooftree}
\qquad
\begin{prooftree}
\varphi \leftrightarrow \varphi'
\justifies (\psi \Rightarrow \varphi) \leftrightarrow (\psi \Rightarrow \varphi')  \using (RCEC)
\end{prooftree}
\end{center}

We refer to $CK$ \cite{Chellas80} as the minimal quantified conditional logic
closed under rules RCEA, RCEC and RCK.  In what follows, only
quantified conditional logics extending CK are considered.

\section{Classical Higher-Order Logic}\label {sec:hol}
HOL is a logic based on simply typed $\lambda$-calculus
\cite{Church40,AndrewsSEP}.  The set $\mathcal{T}$ of simple types in
HOL is usually freely generated from a set of basic types $\{o,i\}$
using the function type
constructor $\typearrow$. Here we instead consider a set of basic type 
$\{o,i,u\}$, where $o$ denotes the type of Booleans, and where $i$ and $u$ denote  
some non-empty domains.  Without loss of generality, we will later identify $i$ with a 
 set of worlds and $u$ with a domain of individuals.

Let $\alpha,\beta,o \in \mathcal{T}$. The \emph{terms} of HOL are defined by 
the grammar ($p_{\alpha}$ denotes typed constants and $X_{\alpha}$ typed variables
distinct from $p_\alpha$):
$$s,t ::=  p_{\alpha} \mid X_{\alpha} \mid (\lambda X_{\alpha}.s_{\beta})_{\alpha \typearrow \beta} \mid 
(s_{\alpha \typearrow \beta})_{\beta} \mid (\neg_{o \typearrow o}\;s_{o})_o \mid 
 (s_o \vee_{o\typearrow o \typearrow o} t_o)_o 
\mid (\Pi_{(\alpha \typearrow o)\typearrow o}\; s_{\alpha \typearrow o})_o$$

Complex typed terms are constructed via abstraction and
application. The primitive logical connectives are $\neg_{o \typearrow
  o},\vee_{o \typearrow o \typearrow o}$ and $\Pi_{(\alpha \typearrow
  o) \typearrow o}$ (for each type $\alpha$). From these, other
logical connectives can be introduced as abbreviations: for example, $\wedge$
and $\rightarrow$ abbreviate the terms $\lambda A. \lambda B. \neg
(\neg A \vee \neg B)$ and $\lambda A. \lambda B. \neg A \vee B$,
etc. HOL terms of type $o$ are called formulas.
Binder notation $\forall X_{\alpha}.s_o$ is used as an abbreviation
for ($\Pi_{(\alpha \typearrow o)\typearrow o}\;(\lambda
X_{\alpha}. s_{o})$).  Substitution of a term $A_{\alpha}$ for a
variable $X_{\alpha}$ in a term $B_{\beta}$ is denoted by $[A/X]B$, where it is
assumed that the bound variables of $B$ avoid variable capture. Well known operations and relations on HOL terms 
include $\beta\eta$-normalization and $\beta\eta$-equality, denoted by  $s =_{\beta\eta} t$. 

The following definition of HOL semantics 
closely follows the standard literature \cite{Andrews1972b,AndrewsSEP}.

A \emph{frame} is a collection $\{D_{\alpha}\}_{\alpha \in
  \mathcal{T}}$ of nonempty sets called \emph{domains} such that
$D_{o} = \{T,F\}$ where $T$ represents truth and $F$ falsehood,
$D_{i}\not=\emptyset$ and $D_{u}\not=\emptyset$ are chosen arbitrary,
and $D_{\alpha \typearrow \beta}$ are collections of total functions mapping
$D_{\alpha}$ into $D_{\beta}$.

  An \emph{interpretation} is a tuple $\langle \{D_{\alpha}\}_{\alpha \in
    \mathcal{T}}, I \rangle$ where $\{D_{\alpha}\}_{\alpha \in
    \mathcal{T}}$ is a frame and where function $I$ maps each typed
  constant $c_{\alpha}$ to an appropriate element of $D_{\alpha}$,
  which is called the \emph{denotation} of $c_{\alpha}$. The
  denotations of $\neg,\vee$ and $\Pi_{(\alpha \typearrow o)\typearrow
    o}$ are always chosen as usual. A variable assignment $\phi$ maps
  variables $X_{\alpha}$ to elements in $D_{\alpha}$. 

  An interpretation is a \emph{Henkin model (general model)} if and
  only if there is a binary valuation function $\mathcal{V}$ such that
  $\mathcal{V}(\phi,s_{\alpha}) \in D_{\alpha}$ for each variable
  assignment $\phi$ and term $s_{\alpha}$, and the following
  conditions are satisfied for all $\phi$, variables $X_\alpha$,
  constants $p_\alpha$, and terms $l_{\alpha \typearrow \beta},
  r_\alpha, s_\beta$ (for $\alpha,\beta\in\mathcal{T}$):
  $\mathcal{V}(\phi,X_{\alpha}) = \phi(X_\alpha)$,
  $\mathcal{V}(\phi,p_{\alpha}) = I(p_{\alpha})$,
  $\mathcal{V}(\phi,(l_{\alpha \typearrow \beta}\; r_{\alpha})) =
  (\mathcal{V}(\phi,l_{\alpha \typearrow
    \beta}))(\mathcal{V}(\phi,r_{\alpha}))$, and
  $\mathcal{V}(\phi,\lambda X_{\alpha}.s_{\beta})$ represents the
  function from $D_{\alpha}$ into $D_{\beta}$ whose value for each
  argument $z \in D_{\alpha}$ is $\mathcal{V}(\phi[z/X_{\alpha}],
  s_\beta)$, where $\phi[z/X_\alpha]$ is that variable assignment such
  that $\phi[z/X_\alpha](X_\alpha) = z$ and $\phi[z/X_\alpha]Y_\beta =
  \phi Y_\beta$ when $Y_\beta \not=X_\alpha$.

If an interpretation $\mathcal{H} = \langle \{D_\alpha\}_{\alpha \in
  \mathcal{T}}, I\rangle$ is an \emph{Henkin model} the function
$\mathcal{V}$ is uniquely determined and
$\mathcal{V}(\phi,s_\alpha)\in D_\alpha$ is called the denotation of
$s_\alpha$. $\mathcal{H}$ is called a \emph{standard model} if and
only if for all $\alpha$ and $\beta$, $D_{\alpha \typearrow \beta}$ is
the set of all functions from $D_{\alpha}$ into $D_{\beta}$. It is
easy to verify that each standard model is also a Henkin model. A
formula $A$ of HOL is \emph{valid} in a Henkin model $\mathcal{H}$ if
and only if $\mathcal{V}(\phi,A) = T$ for all variable assignments
$\phi$. In this case we write $\mathcal{H}\models A$.  $A$ is (Henkin) valid,
denoted as $\models A$, if and only if $\mathcal{H}\models A$ for all 
Henkin models 
$\mathcal{H}$.

\begin{proposition}\label{prop:trivial}
  Let $\mathcal{V}$ be the valuation function of Henkin model
  $\mathcal{H}$. The following properties hold for all 
  assignments $\phi$, terms $s_o,t_o,l_\alpha,r_\alpha$, and variables
  $X_\alpha,V_\alpha$ (for $\alpha\in\mathcal{T}$): $\mathcal{V}(\phi,(\neg s_o))=T$ if and only if $\mathcal{V}(\phi,s_o)=F$, $\mathcal{V}(\phi,(s_o \vee t_o))=T$ if and only if $\mathcal{V}(\phi,s_o)=T$ or $\mathcal{V}(\phi,s_o)=T$, $\mathcal{V}(\phi,(s_o \wedge t_o))=T$ if and only if $\mathcal{V}(\phi,s_o)=T$ and $\mathcal{V}(\phi,s_o)=T$,  $\mathcal{V}(\phi,(s_o \rightarrow t_o))=T$ if and only if $\mathcal{V}(\phi,s_o)=F$ or $\mathcal{V}(\phi,s_o)=T$,  $\mathcal{V}(\phi,(\forall X_\alpha. s_o))=
  \mathcal{V}(\phi,(\Pi_{(\alpha\typearrow o) \typearrow o}\; (\lambda
  X_\alpha. s_o)))=T$ if and only if  for all $v\in D_\alpha$ holds $\mathcal{V}(\phi[v/V_\alpha],((\lambda
  X_\alpha. s_o)\;V))=T$, and if $l_\alpha =_{\beta\eta} r_\alpha$ then $\mathcal{V}(\phi,l_\alpha)=\mathcal{V}(\phi,r_\alpha)$
\end{proposition}

\section{Embedding Quantified Conditional Logics in HOL}\label{sec:embedding}

Quantified conditional logic formulas are identified with certain HOL
terms (predicates) of type $i \typearrow o$.  They can be applied to
terms of type $i$, which are assumed to denote possible worlds. 

\begin{definition}\label{def:embedding}
  The mapping $\lfloor \cdot \rfloor$ translates
  formulas $\varphi$ of quantified conditional logic $CK$ into HOL terms $\lfloor
  \varphi \rfloor$ of type $i \typearrow o$. The mapping is recursively defined as follows:
\[
\begin{array}{lcl}
\lfloor P \rfloor &=& P_{i \typearrow o} \\
\lfloor k(X^1,\ldots,X^n) \rfloor &=& (\lfloor k \rfloor \lfloor X^1 \rfloor \ldots \lfloor X^n \rfloor) \rfloor \\
                             &=& (k_{u^n \typearrow (i \typearrow o)}\;X^1_u \ldots X^n_u)\\
\lfloor \neg \varphi \rfloor &=&  \neg_{i \typearrow o}\;\lfloor \varphi \rfloor \\
\end{array} \qquad \qquad 
\begin{array}{lcl}
\lfloor \varphi\vee \psi \rfloor &=&  \vee_{(i \typearrow o) \typearrow (i \typearrow o) \typearrow (i \typearrow o)}\;\lfloor\varphi \rfloor \lfloor \psi \rfloor\\
\lfloor \varphi\Rightarrow \psi \rfloor &=&  \Rightarrow_{(i \typearrow o) \typearrow (i \typearrow o) \typearrow (i \typearrow o)}\;\lfloor\varphi \rfloor \lfloor \psi \rfloor \\
\lfloor \forall X. \varphi \rfloor &=&   \Pi_{(u\typearrow(i \typearrow o))\typearrow(i\typearrow o)}\;\lambda X_u. \lfloor\varphi \rfloor \\
\lfloor \forall P. \varphi \rfloor &=&  \Pi_{((i\typearrow o)\typearrow(i \typearrow o))\typearrow(i \typearrow o)}\;\lambda P_{i \typearrow o}. \lfloor\varphi \rfloor
\end{array}
\]
$P_{i \typearrow o}$ and $X^1_u,\ldots,X^n_u$ are HOL variables and
$k_{u^n \typearrow (i \typearrow o)}$ is a HOL
constant.  $\neg_{i \typearrow o}$, $\vee_{(i \typearrow o) \typearrow
  (i \typearrow o) \typearrow (i \typearrow o)}$, $\Rightarrow_{(i
  \typearrow o) \typearrow (i \typearrow o) \typearrow (i \typearrow
  o)}$, $\Pi_{(u\typearrow(i \typearrow o))\typearrow(i\typearrow o)}$
and $\Pi_{((i\typearrow o)\typearrow(i \typearrow o))\typearrow(i
  \typearrow o)}$ realize the quantified conditional logics connectives in
HOL. They abbreviate the following proper HOL terms:
\[
\begin{array}{ll}
\neg_{(i \typearrow o) \typearrow (i \typearrow o)} &= \lambda A_{i \typearrow o}.\lambda X_{i}. \neg(A\, X)\\
\vee_{(i \typearrow o) \typearrow (i \typearrow o) \typearrow (i \typearrow o)} &= \lambda A_{i \typearrow o}.\lambda B_{i \typearrow o}.\lambda X_i. (A\,X) \vee (B\,X)\\
\Rightarrow_{(i \typearrow o) \typearrow (i \typearrow o) \typearrow (i \typearrow o)} &= \lambda A_{i \typearrow o}.\lambda B_{i \typearrow o}.\lambda X_{i}.\forall W_i.(f\, X\, A\, W) \rightarrow (B\, W)\\
\Pi_{(u\typearrow(i \typearrow o))\typearrow(i \typearrow o)} & = \lambda {Q_{u\typearrow(i \typearrow o)}}. \lambda {W_i}. \forall{X_u}. (Q\,X\,W) \\
\Pi_{((i \typearrow o)\typearrow(i \typearrow o))\typearrow(i \typearrow o)} & = \lambda {R_{(i \typearrow o)\typearrow(i \typearrow o)}}. \lambda {W_i}. \forall {P_{i\typearrow o}}. (R\,P\,W)
\end{array}
\]
The constant symbol $f$ in the mapping of $\Rightarrow$ is of type ${i
  \typearrow (i \typearrow o) \typearrow (i \typearrow o)}$. It
realizes the selection function, i.e., its interpretation is chosen
appropriately (cf. below).

This mapping induces mappings $\lfloor\mathcal{IV}\rfloor$,
$\lfloor\mathcal{PV}\rfloor$ and $\lfloor\mathcal{SYM}\rfloor$ of the
sets $\mathcal{IV}$, $\mathcal{PV}$ and $\mathcal{SYM}$ respectively.
\end{definition}

Analyzing the validity of a translated formula $\lfloor \varphi
\rfloor$ for a world represented by term $t_i$ corresponds to
evaluating the application $(\lfloor \varphi \rfloor\;t_i)$.  In line
with \cite{J21}, we define
$ \text{vld}_{(i \typearrow o)\typearrow o} = \lambda A_{i \typearrow o}. \forall S_{i}.(A\; S)$.
With this definition, validity of a quantified conditional formula $\varphi$ in
CK corresponds to the validity of the corresponding formula $
(\text{vld}\; \lfloor \varphi \rfloor)$ in HOL, and vice versa.

\section{Soundness and Completeness}
To prove the soundness and completeness of the embedding, a
mapping from selection function models into Henkin models is employed. 
This mapping will employ a corresponding mapping of variable assignments for
quantified conditional logics into variable assignments for HOL.

\begin{definition}[Mapping of Variable Assignments]
Let $g = (g^{iv}:\mathcal{IV}\longrightarrow D,\, g^{pv}:\mathcal{PV}\longrightarrow
Q)$ be a variable assignment for a quantified conditional logic.  We define the corresponding variable assignment
$\lfloor g \rfloor=(\lfloor g^{iv}\rfloor:\lfloor \mathcal{IV} \rfloor\longrightarrow
D,\;\lfloor g^{pv}\rfloor:\lfloor \mathcal{PV} \rfloor\longrightarrow Q)$ for HOL so that $\lfloor g \rfloor(X_u) =
\lfloor g \rfloor(\lfloor X \rfloor) = g(X)$ and $\lfloor g \rfloor(P_{i\typearrow o}) =
\lfloor g \rfloor(\lfloor P\rfloor) = g(P)$ for all $X_u\in\lfloor \mathcal{IV} \rfloor$ and
$P_{i\typearrow o}\in\lfloor \mathcal{PV} \rfloor$.
Finally, a variable assignment $\lfloor g \rfloor$ is extended to an assignment for variables $Z_\alpha$ of arbitrary type by choosing $\lfloor g \rfloor(Z_\alpha) = d\in D_\alpha$ arbitrary, if $\alpha\not= u,{i\typearrow o}$.
\end{definition}

\begin{definition}[Henkin model $\mathcal{H}^{\mathcal{M}}$]
\label{embedding}
  Given a quantified conditional logic model  $\mathcal{M} = \langle S,f,D,Q, I \rangle$. The Henkin model 
  $\mathcal{H}^{\mathcal{M}} = \langle \{D_\alpha\}_{\alpha \in
    \mathcal{T}}, I \rangle$ for $\mathcal{M}$ is defined as follows: $D_i$ is chosen as the set of possible worlds $S$,  $D_u$ is chosen as the first-order domain $D$ (cf.\ definition of $\lfloor g^{iv} \rfloor$),  $D_{i\typearrow o}$ is chosen as the set of sets of possible worlds $Q$ (cf.\ definition of $\lfloor g^{pv} \rfloor$)\footnote{To keep things simple, we identify sets with their characteristic functions.}, and  all other sets $D_{\alpha\typearrow\beta}$ are chosen as (not
  necessarily full) sets of functions from $D_\alpha$ to $D_\beta$.
  For all sets $D_{\alpha\typearrow\beta}$ the rule that everything
  denotes must be obeyed, in particular, we require that the sets
  $D_{u^n\typearrow(i\typearrow o)}$ and $D_{i\typearrow(i\typearrow o)\typearrow(i\typearrow o)}$ contain the elements $I
  k_{u^n\typearrow(i\typearrow o)}$ and $I
  f_{i\typearrow(i\typearrow o)\typearrow(i\typearrow o)}$ as characterized below.

\sloppy The interpretation $I$ is constructed as follows: 
(i) Let $k_{u^n\typearrow(i\typearrow o)} = \lfloor k \rfloor$ for
  $n$-ary $k\in\mathcal{SYM}$ and let $X^i_u = \lfloor X^i \rfloor$
  for $X^i\in\mathcal{IV}$, $i=1,\ldots,n$.  We choose $I
  k_{u^n\typearrow(i\typearrow o)} \in D_{u^n\typearrow(i\typearrow
    o)}$ such that $(I\,k_{u^n\typearrow(i\typearrow o)})(\lfloor g \rfloor(X^1_u),\ldots,\lfloor g \rfloor(X^n_u),w) = T$ for all
  worlds $w\in D_i$ such that $\mathcal{M},g,w\models k(X^1,\ldots,X^n)$, that
  is, if $\langle g^{iv}(X^1),\ldots,g^{iv}(X^n)\rangle\in I(k,w)$.   Otherwise we choose $(I\,k_{u^n\typearrow(i\typearrow o)})
  (\lfloor g \rfloor(X^1_u),\ldots,\lfloor g \rfloor(X^n_u),w) = F$.
%
(ii)  We choose $I f_{i \typearrow (i \typearrow o) \typearrow (i
    \typearrow o)} \in D_{i \typearrow (i \typearrow o) \typearrow (i
    \typearrow o)}$ such that 
$(I f_{i \typearrow (i \typearrow o)
  \typearrow (i \typearrow o)})(s,q,t) = T$ for all worlds $s,t\in
D_i$ and $q\in D_{i \typearrow o}$ with $t \in f(s,\{x\in S\mid q(x) =
T\})$ in $\mathcal{M}$. Otherwise we choose $(I f_{i \typearrow (i \typearrow o)
  \typearrow (i \typearrow o)})(s,q,t) = F$.
(iii)  For all other constants $s_\alpha$, choose $I s_\alpha$ arbitrary.\footnote{In fact, we may safely assume that there are
  no other typed constant symbols given, except for the symbol $f_{i
    \typearrow (i \typearrow o) \typearrow (i \typearrow o)}$, the
  symbols $,k_{u^n\typearrow(i\typearrow o)}$, and the logical connectives.}

It is not hard to verify that $\mathcal{H}^{\mathcal{M}}$ is a Henkin
model.
\end{definition}

\begin{lemma}\label{lemma:1} \sloppy
  Let $\mathcal{H}^{\mathcal{M}}$ be a Henkin model for a selection
  function model $\mathcal{M}$. For all quantified conditional logic formulas
  $\delta$, variable assignments $g$ and worlds $s$ it
  holds: $\mathcal{M},g,s \models \delta \text{\ if and only if\ }
  \mathcal{V}(\lfloor g \rfloor [s/S_i], (\lfloor \delta \rfloor\; S_i))=T$
\begin{proof} 
  The proof is by induction on the structure of $\delta$.
  The cases for $\delta = P$, $\delta = k(X^1,\ldots,X^n)$, $\delta =
  (\neg \varphi)$, $\delta = (\varphi \vee \psi)$, and $\delta =
  (\varphi \Rightarrow \psi)$ are similar to Lemma 1 in \cite{R52}.
  The cases for $\delta = \forall {X}. \varphi$ and $\delta = \forall
  {P}. \varphi$ adapt the respective cases from Lemmas 4.3 and 4.7 in
  \cite{J23}.

\end{proof}
\end{lemma}

\begin{theorem}[Soundness and Completeness]
  $\models (\text{vld}\; \lfloor \varphi \rfloor) \text{ in HOL if and
  only if } \models \varphi \text{ in CK}$
\begin{proof} (Soundness) The proof is by contraposition. Assume
  $\not\models \varphi$ in CK, that is, there is a model $\mathcal{M}
  = \langle S,f,D,Q,I \rangle$, a variable assignment $g$ and a world
  $s\in S$, such that $\mathcal{M},g,s\not\models \varphi$. By Lemma
  \ref{lemma:1} we have that $\mathcal{V}(\lfloor g \rfloor
  [s/S_i],(\lfloor \varphi \rfloor\;S)) = F$ in Henkin model
  $\mathcal{H}^\mathcal{M}= \langle \{D_\alpha\}_{\alpha \in
    \mathcal{T}}, I \rangle$ for $\mathcal{M}$. Thus, by
  Prop.~\ref{prop:trivial}, definition of $\text{vld}$ and since $(\forall
  S_i. \lfloor \varphi \rfloor\;S) =_{\beta\eta} (\text{vld}\;\lfloor
  \varphi \rfloor)$ we know that $\mathcal{V}(\lfloor g
  \rfloor,(\forall S_i. \lfloor \varphi \rfloor\;S)) =
  \mathcal{V}(\lfloor g \rfloor,(\text{vld}\;\lfloor \varphi \rfloor))
  = F$. Hence, $\mathcal{H}^\mathcal{M}\not\models
  (\text{vld}\;\lfloor \varphi \rfloor)$, and thus $\not\models
  (\text{vld}\;\lfloor \varphi \rfloor)$ in HOL.

  (Completeness) The proof is again by contraposition. Assume
  $\not\models (\text{vld}\; \lfloor \varphi \rfloor)$ in HOL, that
  is, there is a Henkin model $\mathcal{H}=\langle
  \{D_\alpha\}_{\alpha \in \mathcal{T}}, I \rangle$ and a variable
  assignment $\phi$ with $\mathcal{V}(\phi,(\text{vld}\; \lfloor
  \varphi \rfloor)) = F$.  Without loss of generality we can assume
  that Henkin Model $\mathcal{H}$ is in fact a Henkin model
  $\mathcal{H}^\mathcal{M}$ for a corresponding quantified conditional
  logic model $\mathcal{M}$ and that $\Phi = \lfloor g \rfloor$ for a
  corresponding quantified conditional logic variable assignment
  $g$.
  By Prop.~\ref{prop:trivial}  and since $(\text{vld}\;\lfloor
  \varphi \rfloor) =_{\beta\eta} (\forall S_i. \lfloor \varphi
  \rfloor\;S)$ we have $\mathcal{V}(\lfloor g \rfloor,(\forall
  S_i. \lfloor \varphi \rfloor\;S)) = F$, and hence, by definition of
  $\text{vld}$, $\mathcal{V}(\lfloor g \rfloor [s/S_i],\lfloor \varphi
  \rfloor\;S) = F$ for some $s\in D$. By Lemma \ref{lemma:1} we thus
  know that $\mathcal{M},g,s \not \models \varphi$, and hence
  $\not\models \varphi$ in CK.
\end{proof}
\end{theorem} 

\section{Conclusion} We have presented an embedding of quantified
conditional logics in HOL. This embedding enables the uniform
application of higher-order automated theorem provers and model
finders for reasoning about and within quantified conditional
logics. In previous work we have studied related embeddings in HOL,
including propositional conditional logics \cite{R52} and quantified
multimodal logics \cite{J23}. First experiments with these embeddings
have provided evidence for their practical relevance. Moreover, an
independent case study on reasoning in quantified modal logics shows
that the embeddings based approach may even outperform specialist
reasoners quantified modal logics
\cite{jens11:_implem_and_evaluat_theor_prover}. Future work will
investigate whether HOL reasoners perform similarly well also for
quantified conditional logics. For a first impression of such studies
we refer to the Appendices \ref{app:embedding} and \ref{app:ex1},
where we also present the concrete encoding of our embedding in TPTP
THF0 \cite{J22} syntax. Unfortunately we are not aware of any other
(direct or indirect) prover for quantified conditional logics that
could be used for comparison.

\bibliographystyle{plain}

\begin{appendix}
\section{The Embedding of Quantified Conditional Logic in HOL in THF0 Syntax} \label{app:embedding}
We present an encoding of our embedding of quantified conditional logics in HOL in the  TPTP THF0 \cite{J22} syntax. 

Satisfiability of this embedding is shown by the HOL reasoner Satallax\footnote{\url{http://www.ps.uni-saarland.de/~cebrown/satallax/}} in only 0.01 seconds.

{\small
\begin{verbatim}
%---------------------------------------------------------------------
%---- reserved constant for selection function f
thf(f_type,type,(
    f: $i > ( $i > $o ) > $i > $o )).

%---- 'not' in conditional logic
thf(cnot_type,type,(
    cnot: ( $i > $o ) > $i > $o )).

thf(cnot_def,definition,
    ( cnot
    = ( ^ [Phi: $i > $o,X: $i] :
          ~ ( Phi @ X ) ) )).

%---- 'or' in conditional logic
thf(cor_type,type,(
    cor: ( $i > $o ) > ( $i > $o ) > $i > $o )).

thf(cor_def,definition,
    ( cor
    = ( ^ [Phi: $i > $o,Psi: $i > $o,X: $i] :
          ( ( Phi @ X )
          | ( Psi @ X ) ) ) )).

%---- 'true' in conditional logic
thf(ctrue_type,type,(
    ctrue: $i > $o )).

thf(ctrue_def,definition,
    ( ctrue
    = ( ^ [X: $i] : $true ) )).

%---- 'false' in conditional logic
thf(cfalse_type,type,(
    cfalse: $i > $o )).

thf(cfalse_def,definition,
    ( cfalse
    = ( ^ [X: $i] : $false ) )).

%---- 'conditional implication' in conditional logic
thf(ccond_type,type,(
    ccond: ( $i > $o ) > ( $i > $o ) > $i > $o )).

thf(ccond_def,definition,
    ( ccond
    = ( ^ [Phi: $i > $o,Psi: $i > $o,X: $i] :
        ! [W: $i] :
          ( ( f @ X @ Phi @ W )
         => ( Psi @ W ) ) ) )).

%---- 'and' in conditional logic
thf(cand_type,type,(
    cand: ( $i > $o ) > ( $i > $o ) > $i > $o )).

thf(cand_def,definition,
    ( cand
    = ( ^ [Phi: $i > $o,Psi: $i > $o,X: $i] :
          ( ( Phi @ X )
          & ( Psi @ X ) ) ) )).

%---- 'conditional equivalence' in conditional logic
thf(ccondequiv_type,type,(
    ccondequiv: ( $i > $o ) > ( $i > $o ) > $i > $o )).

thf(ccondequiv_def,definition,
    ( ccondequiv
    = ( ^ [Phi: $i > $o,Psi: $i > $o] :
          ( cand @ ( ccond @ Phi @ Psi ) @ ( ccond @ Psi @ Phi ) ) ) )).

%---- 'material implication' in conditional logic
thf(cimpl_type,type,(
    cimpl: ( $i > $o ) > ( $i > $o ) > $i > $o )).

thf(cimpl_def,definition,
    ( cimpl
    = ( ^ [Phi: $i > $o,Psi: $i > $o,X: $i] :
          ( ( Phi @ X )
         => ( Psi @ X ) ) ) )).

%---- 'material equivalence' in conditional logic
thf(cequiv_type,type,(
    cequiv: ( $i > $o ) > ( $i > $o ) > $i > $o )).

thf(cequiv_def,definition,
    ( cequiv
    = ( ^ [Phi: $i > $o,Psi: $i > $o] :
          ( cand @ ( cimpl @ Phi @ Psi ) @ ( cimpl @ Psi @ Phi ) ) ) )).

%---- 'universal quantification (individuals)' in conditional logic
thf(cforall_ind_type,type,(
    cforall_ind: ( mu > $i > $o ) > $i > $o )).

thf(cforall_ind,definition,
    ( cforall_ind
    = ( ^ [Phi: mu > $i > $o,W: $i] :
        ! [X: mu] :
          ( Phi @ X @ W ) ) )).

%---- 'universal quantification (propositions)' in conditional logic
thf(cforall_prop_type,type,(
    cforall_prop: ( ( $i > $o ) > $i > $o ) > $i > $o )).

thf(cforall_prop,definition,
    ( cforall_prop
    = ( ^ [Phi: ( $i > $o ) > $i > $o,W: $i] :
        ! [P: $i > $o] :
          ( Phi @ P @ W ) ) )).

%---- 'existential quantification (individuals)' in conditional logic
thf(cexists_ind_type,type,(
    cexists_ind: ( mu > $i > $o ) > $i > $o )).

thf(cexists_ind,definition,
    ( cexists_ind
    = ( ^ [Phi: mu > $i > $o] :
          ( cnot
          @ ( cforall_ind
            @ ^ [X: mu] :
                ( cnot @ ( Phi @ X ) ) ) ) ) )).

%---- 'existential quantification (propositions)' in conditional logic
thf(cexists_prop_type,type,(
    cexists_prop: ( ( $i > $o ) > $i > $o ) > $i > $o )).

thf(cexists_prop,definition,
    ( cexists_prop
    = ( ^ [Phi: ( $i > $o ) > $i > $o] :
          ( cnot
          @ ( cforall_prop
            @ ^ [P: $i > $o] :
                ( cnot @ ( Phi @ P ) ) ) ) ) )).

%---- 'validity' of a conditional logic formula
thf(valid_type,type,(
    valid: ( $i > $o ) > $o )).

thf(valid_def,definition,
    ( valid
    = ( ^ [Phi: $i > $o] :
        ! [S: $i] :
          ( Phi @ S ) ) )).
%---------------------------------------------------------------------
\end{verbatim}
}

\section{The Barcan Formula and the Converse Barcan Formula} \label{app:ex1}
Using the above THF0 encoding, the Barcan formula 
$ (\forall X. A \Rightarrow B(x)) \rightarrow (A \Rightarrow \forall X.B(x)) $
can be encoded in THF0 as given below. The HOL provers LEO-II\footnote{http://www.leoprover.org} and Satallax
can both prove this theorem in 0.01 seconds. This confirms that our
encoding assumes constant domain semantics.

{\small
\begin{verbatim}
%---------------------------------------------------------------------
include('CK_axioms.ax').

%---- conjecture statement
thf(a,type,(
    a: $i > $o )).

thf(b,type,(
    b: mu > $i > $o )).

thf(bf,conjecture,
    ( valid
    @ ( cimpl
      @ ( cforall_ind
        @ ^ [X: mu] :
            ( ccond @ a @ ( b @ X ) ) )
      @ ( ccond @ a
        @ ( cforall_ind
          @ ^ [X: mu] :
              ( b @ X ) ) ) ) )).
%---------------------------------------------------------------------    
\end{verbatim}
}

The converse Barcan formula
$ (A \Rightarrow \forall X.B(x)) \rightarrow (\forall X. A \Rightarrow B(x)) $
can be encoded analogously. Again, the HOL provers LEO-II and Stallax need only 0.01 seconds to prove this theorem.

{\small
\begin{verbatim}
%---------------------------------------------------------------------
include('CK_axioms.ax').

%---- conjecture statement
thf(a,type,(
    a: $i > $o )).

thf(b,type,(
    b: mu > $i > $o )).

thf(cbf,conjecture,
    ( valid
    @ ( cimpl
      @ ( ccond @ a
        @ ( cforall_ind
          @ ^ [X: mu] :
              ( b @ X ) ) )
      @ ( cforall_ind
        @ ^ [X: mu] :
            ( ccond @ a @ ( b @ X ) ) ) ) )).
%---------------------------------------------------------------------
\end{verbatim}
}
\end{appendix}

\end{document}